\documentclass[sigconf]{acmart}

\setcopyright{rightsretained}
\copyrightyear{2024}
\acmDOI{}

\acmConference[HRI-Trust '24]{Workshop on Trust in HRI}{March 11, 2024}{Boulder, CO}
\acmISBN{}

\begin{document}

\title{Towards a Participatory and Social Justice-Oriented Measure of Human-Robot Trust}

\author{Raj Korpan}
\email{raj.korpan@hunter.cuny.edu}
\orcid{0000-0003-0431-9134}
\affiliation{%
  \institution{Hunter College, City University of New York}
  \streetaddress{695 Park Ave}
  \city{New York}
  \state{New York}
  \country{USA}
  \postcode{10065}
}

\renewcommand{\shortauthors}{Raj Korpan}

\begin{abstract}
Many measures of human-robot trust have proliferated across the HRI research literature because each attempts to capture the factors that impact trust despite its many dimensions. None of the previous trust measures, however, address the systems of inequity and structures of power present in HRI research or attempt to counteract the systematic biases and potential harms caused by HRI systems. This position paper proposes a participatory and social justice-oriented approach for the design and evaluation of a trust measure. This proposed process would iteratively co-design the trust measure with the community for whom the HRI system is being created. The process would prioritize that community's needs and unique circumstances to produce a trust measure that accurately reflects the factors that impact their trust in a robot.
\end{abstract}





\maketitle

\section{Introduction}\label{sec:introduction}
In many Human-Robot Interaction (HRI) studies the most important evaluation metric is often trust in the robot \cite{law2021trust} because it is believed that trust is the channel through which robots will be accepted and successfully deployed in real-world contexts \cite{kok2020trust}. A human's trust in a robot, however, cannot be directly observed \cite{kok2020trust}. So there is no objective way to directly measure trust during interactions \cite{schaefer2013perception} \footnote{Although some use certain observable behaviors during an interaction as a proxy objective measure, such as compliance with the robots suggestions \cite{salem2015would} or attention focused on the robot as measured by eye tracking \cite{jenkins2010measuring}, these behaviors do not directly observe a human's mental state \cite{kohn2021measurement}.}. Furthermore, it has been suggested that trust is multi-faceted \cite{knowles2022many, khalid2021determinants}, dynamic \cite{kaplan2021time}, subjective \cite{schaefer2016measuring}, and affected by the robot's physical form \cite{bernotat2021fe, fischer2023effect} and performance \cite{hancock2011meta}. These characteristics (lack of observability and complexity) have resulted in the proliferation of many different measures of trust in the HRI community, each focused on different aspects of trust \cite{chita2021can, khavas2021review}. This position paper argues that the development and evaluation of a reliable measure of human-robot trust should be participatory and justice-oriented. 

\section{Prior Work in Measure Development} \label{sec:relatedwork}
Measures of trust are often a self-report questionnaire that consists of a series of Likert-style questions that ask about different aspects of trust \cite{schaefer2016measuring, baker2018toward}. The questions attempt to represent a ``universal'' experience of trust that applies in all situations and contexts (e.g., \cite{muir1990operators, jian2000foundations, madsen2000measuring}). For example, the Merritt trust scale asks participants to respond on a 5-point Likert rating scale to 6 statements, such as ``I trust the [technology]'' \cite{merritt2011affective}. In contrast, the HRI Trust Scale uses a 7-point Likert rating scale with 37 statements across 5 attributes: team configuration, team process, context, task, and system \cite{yagoda2012you}. The Trust Perception Scale-HRI (TPS-HRI) consists of 40 questions all proceeded by ``What percentage of the time will this robot...'' and participants could select between 0\% and 100\% in increments of 10\% \cite{schaefer2016measuring}. The Multi-Dimensional Measure of Trust (MDMT) measures trust with 20 items across five dimensions: reliability, competency, ethicality, transparency and benevolence \cite{ullman2019measuring, malle2021multidimensional}.

The differences in these measures demonstrate the disparate ways in which they have been developed and evaluated. For example, the Merritt trust scale was created by the author based on their expertise and then only evaluated to show that trust was related to but distinct from the propensity to trust machines or like them \cite{merritt2011affective}. This approach is susceptible to encoding the individual biases of the author. It was also not validated for its generalizability.

Yagoda and Gillan's HRI Trust Scale started with a preliminary list of items, conducted an exploratory study where 11 HRI subject matter experts (SMEs) provided feedback on this initial list, generated a list of HRI trust scale items based on the feedback, conducted an online crowdsourced quality assessment with 100 participants on Amazon Mechanical Turk (MTurk), and then used factor analysis to determine the final HRI Trust Scale \cite{yagoda2012you}. Although this approach is an improvement over an individual's independent creation of a scale, it still has several limitations. First, the HRI SMEs used to provide feedback were required to have at least 5 years of professional experience and have published contributions in the field of HRI \cite{yagoda2012you}. The authors do not mention any other demographic information about these experts, so there is no way to know what biases they may have introduced in their feedback. Second, the use of these experts reinforces the structures of power in the robotics community \cite{williams2024understanding}. Reliance on these experts maintains the historical inequities of who robotic systems are built by and for because the robotics community has often excluded Black people \cite{howard2020robots}, women \cite{graesser2021gender}, queer people \cite{OrganizersOfQueerin2023QueerAI,korpan2023trust}, and other underrepresented groups \cite{tanevska2023inclusive}.

The TPS-HRI began with an initial list of 156 items identified from a literature review of over 700 papers, which contained 86 trust scales, and two initial experiments to identify physical attributes that affect trustworthiness \cite{schaefer2016measuring}. Each item on this initial list was evaluated on a 7-point Likert rating scale by 159 undergraduate students and statistical analysis (e.g., principal component analysis) was used to reduce the number of items to 73 \cite{schaefer2016measuring}. The scale was also changed to use percentage increments at this point so that participants rate items on a range from no trust (0\%) to complete trust (100\%) \cite{schaefer2016measuring}. Next, 11 SMEs were surveyed to evaluate each item's importance to include in a trust scale and their feedback resulted in the reduction of items to 42 \cite{schaefer2016measuring}. These SMEs were recruited from the United States Army and Air Force Research Laboratories and university research laboratories and had 4 to 30 years of experience in robotics research \cite{schaefer2016measuring}.  Similar to the HRI Trust Scale, the use of SMEs can be problematic for the same reasons. Finally, two studies were conducted with undergraduate students to evaluate the validity of the 42-item scale to see if it was able to capture the dynamics of trust over time and whether it actually measures trust \cite{schaefer2016measuring}. The results of these final studies eliminated 2 items from the scale and showed that the final 40 items successfully measured trust across multiple interactions \cite{schaefer2016measuring}.

Although the final TPS-HRI is validated, it mainly relied on convenience samples of undergraduate students with only two demographic characteristics reported: binary gender (male or female) and mean age (for only the first validation study)  \cite{schaefer2016measuring}. This is likely not a representative sample for many of the other applications in which this scale may be applied to measure trust \cite{baxter2016characterising}. Another issue with the validation studies was that it was conducted in simulation in the context of a soldier's interaction with a robot \cite{schaefer2016measuring}. A scale validated to measure trust in this scenario may not necessarily work in other tasks or physical interactions.

The MDMT began with a list of 62 words related to trust collected from dictionaries, related literature, and other trust scales \cite{malle2021multidimensional}. An initial MTurk study evaluated whether each term related more to capacity trust or personal trust, two dimensions of trust identified from the human-robot trust literature \cite{malle2021multidimensional}. Principal component analysis and clustering were then used to identify a reduced list of 20 items but split into 4 dimensions (reliable, capable, sincere, and ethical) \cite{malle2021multidimensional}. A second MTurk study then asked participants to sort an expanded list of 32 items into the 4 dimensions or an ``other'' category \cite{malle2021multidimensional}. The results were used to select the top three items for each dimension and a face-valid item was added for each, which resulted in the initial scale with 16 items that is evaluated on a 7-point Likert rating scale \cite{malle2021multidimensional}. A MTurk study was then used to validate the sensitivity of each dimension to respond to a change in trust related to that dimension and the internal consistency of the items in each dimension \cite{malle2021multidimensional}. This validation, however, was done where participants were presented with a sentence that described a robot's behavior to evaluate their initial trust ratings, and then new information would be given and trust ratings evaluated again \cite{malle2021multidimensional}. Similar to the TPS-HRI, this approach to validation may not result in a trust measure that reliably works in actual human-robot interactions (simulated or physical).

Subsequent work revised the MDMT to update the dimensions (reliable, competent, ethical, transparent, and a fifth dimension, benevolence) for a total of 20 items, again based on an online study where participants had to sort and group terms, and then validated it in a similar way as the initial MDMT \cite{ullman2021developing, malle2023measuring}. Although an online crowdsourcing tool may produce a more representative sample than a set of undergraduate students, other research has investigated ethical concerns regarding the use of MTurk \cite{moss2023ethical}, shown significant levels of inattentiveness among MTurk workers \cite{saravanos2021hidden}, and explored the many data quality concerns with MTurk workers \cite{hauser2019common}. Furthermore, the representativeness of MTurk samples is also questionable -- in the MDMT validation study, for example, 76.5\% of the sample self-reported as ``Non-Hispanic White or Euro-American'' \cite{ullman2021developing} which is certainly not representative of the world population or the United States population \cite{Census2020DECENNIALDP2020.DP1}.

The development of both the TPS-HRI and the MDMT took a different approach to create their initial list of items compared to the Merritt trust scale and the HRI Trust Scale -- they drew items from the existing literature and trust scales. Although they did not use SMEs to generate this initial list, the power structures and norms are still reinforced in this approach because the previous research that was used was developed by the established robotics community \cite{williams2024understanding, tanqueray2023norms}. The next section proposes an alternative approach for the development and evaluation of a measure of human-robot trust that incorporates principles of participatory design and aligns with a social justice-oriented framework for human-robot interaction.

\section{Proposed Approach}\label{sec:proposedapproach}
Because robots have the opportunity to shape our society, their designers can reinforce or disrupt systems of inequity \cite{vsabanovic2010robots} and force their cultural norms on the communities where these technologies are deployed \cite{hipolito2023enactive}. Beyond interpersonal power, robots can wield structural, disciplinary, and cultural power that can reinforce the structures of White Patriarchy \cite{williams2024understanding}. Recent work has shown how discrimination and bias exist at multiple levels in the robotics and artificial intelligence research communities \cite{fosch2022diversity, fosch2023accounting}. As a result, there have been several calls for a more equitable HRI, such as the Feminist HRI framework \cite{Winkle2023feminist}, the HRI Equitable Design framework \cite{ostrowski2022ethics}, and the Robots for Social Justice framework \cite{zhu2024robots}. Inspired by these calls, this position paper suggests a participatory and social justice-oriented human-robot trust measure.

\subsection{Participatory Design}
Participatory design is a collaborative process that directly involves the people affected by technology in its creation \cite{spinuzzi2005methodology}. Participatory design approaches in HRI have been used to design systems that meet the needs of specific users, such as the elderly \cite{rogers2022maximizing} or people with physical or mental disabilities \cite{lee2017steps, azenkot2016enabling, axelsson2019participatory, arevalo2021reflecting}. Participatory design has also been successfully used to develop autonomous social robots \cite{winkle2021leador, tian2021redesigning, axelsson2021social}. Participatory design, however, has not been used in the development of trust measures. As described in Section \ref{sec:relatedwork}, prior approaches were created based initially on an author's expertise, from SMEs input, or a literature review of past trust scales. Subsequently, some of those were evaluated for their validity with undergraduate students or online MTurk workers. A participatory approach would instead engage individuals in the community for whom a robot system is intended to benefit to help co-design the trust measure.

First, the community members would identify the relevant attributes that could affect their trust in a robot, which would be used to produce an initial set of questions. Next, the community members would use the trust measure to evaluate their trust after a physical interaction task with a robot. A questionnaire would then ask them to evaluate the appropriateness of each question (e.g., on a 3-point Likert rating scale ``Not necessary'', ``Useful, but not essential'', and ``Essential'' \cite{lawshe1975quantitative}). A semi-structured interview would also be used to analyze the trust measure's validity -- do the questions allow the participants to express the level of trust they felt during the interaction? Together, the quantitative and qualitative results would be used to revise the initial measure to reflect the participants' feedback. This process would then be iteratively repeated in a cycle with the participants using the revised measure to evaluate their trust, followed by a quantitative and qualitative examination of their experience used to further revise the measure. The participatory design process would stop when the measure converges to a consistent set of questions that accurately represent the attributes that affect trust in that community.

Only one work used a similar participatory process but to develop an initial understanding of trust in the context of robot assistants for elderly care \cite{schwaninger2021you}. In this work, a card game was designed and used during interviews to elicit trust-related factors from older adults \cite{schwaninger2021you}. The card game allowed researchers to ask about specific trust characteristics but also facilitated open conversation about trust \cite{schwaninger2021you}. The results showed that this methodology allowed them to identify the specific dimensions of trust that were relevant to the community they were investigating, such as privacy, control, companionship, and skepticism \cite{schwaninger2021you}. As the authors of that work state, this research represents the initial ideation phase of participatory design, which corresponds to the initial step of the participatory process proposed here. They did not, however, use the findings to produce a trust measure, as proposed here

Although there are potential risks in the use of participatory design, such as possible physical or mental harm, risk of exploitation, and removal of agency \cite{zytko2022ethical}, there are also potential benefits, such as bringing visibility to marginalized people in HRI, addressing social justice issues, and revealing HRI researchers internal biases \cite{lee2022configuring}. For example, a participatory design process that involves queer people could result in a culturally-competent tool \cite{fine2019critical}, more inclusive language \cite{meyer2022rainbow}, center their experiences \cite{korpan2023trust, stolp2024more}, and ultimately produce a technology that they trust more \cite{haimson2020designing, haimson2023transgender}.

The use of participatory design would also support a recent recommendation to center the human experience of trust in the development of a framework for human-robot trust \cite{lange2023human}. It could allow for the development of ad hoc trust measures that address other circumstances that impact trust: repeated interactions, imperfect interactions, subconscious influences on perception, conformity to social norms, and environmental factors \cite{holthaus2023common}. It could also produce context-based questions that address different individual identity factors that affect trust \cite{korpan2023trust} and how those relate to aspects of the robot, task, environment, and other agents \cite{holthaus2023common}. It could also potentially weigh the importance of trust items differently based on community differences \cite{korpan2023trust}.

\subsection{Social Justice Framework}
Robotics researchers' choices reflect the values and beliefs of society, both the good and bad \cite{michalec2021robotics}. For example, the deployment of language-capable robots poses several potential risks, such as overtrust being manipulated to influence human morals, reinforcement of gendered and racialized biases, and perpetuation of harmful assumptions about human identity \cite{williams2023voice}. HRI studies also often fail to broadly address diversity and inclusion in both the human subjects used and the systems created \cite{Seaborndiversity, winkle202315}. A systematic review of HRI research from 2006 to 2022 found that over 90\% of papers used ``WEIRD'' (Western, Educated, Industrial, Rich, and Democratic) participants \cite{Seaborn2023not}. Furthermore, a robot's stereotypical (gendered) appearance have been shown to activate human stereotypes, which subsequently affects people's behavior with those robots and their trust in them \cite{perugia2023models,perugia2023robot}. As a result of these pervasive challenges in how HRI research is conducted, robots are designed and deployed, and who they are evaluated with, a social justice-oriented approach could result in a more equitable HRI \cite{zhu2024robots}.

The Robots for Social Justice framework recommends five considerations for HRI researchers: what the community the research is intended to benefit, what human capabilities would be enhanced, how those capabilities are valued and prioritized, what structural conditions are present in that community, and what are the power structures that surround and influence the community \cite{zhu2024robots}. A social justice-oriented process for the development of a human-robot trust measure should then be grounded in the context in which it will be used. It should begin with a specification of the community for which trust will be measured, consider what human capabilities are being enhanced by the robot, and prioritize the needs of that community as defined by them. As the community is engaged in this development process, the structural and systematic limitations and the power structures that exist in the community should be critically examined. This would ensure that the developed trust measure considers the potential risks and harms that could result if it fails to capture important trust attributes for that community, given its unique circumstances. A participatory design process could easily integrate with this social justice framework to produce a measure of trust that addresses the social and ethical challenges of HRI research.

\section{Conclusion}
Human trust in a robot is a difficult construct to measure because of its complexity. Many different measurement tools have been created and validated in a variety of ways. None of the past approaches, however, addressed the power structures or systems of inequity that HRI research can reinforce. This position paper proposes the use of a participatory design process in the context of a social justice framework would result in a validated trust measure that captures the relevant factors that affect trust for a community.

\bibliographystyle{ACM-Reference-Format}
\bibliography{root}

\end{document}